\documentclass[graybox,vecarrow]{svmult}

\usepackage{mathptmx}       
\usepackage{helvet}         
\usepackage{courier}        
\usepackage{type1cm}        
\usepackage{makeidx}         
\usepackage{graphicx}        
\usepackage{multicol}        
\usepackage[bottom]{footmisc}

\makeindex             
                       
\usepackage{ifthen}
\usepackage{bm}
\usepackage[all]{xy}    

\usepackage{algorithm}
\usepackage{algpseudocode}
\usepackage{wrapfig}
\usepackage[rightcaption]{sidecap}
\sidecaptionvpos{figure}{c}

\usepackage[square,numbers]{natbib}
\usepackage{enumitem}
\newcommand{\eref}[1]{(\ref{#1})}
\newcommand{\secref}[1]{Section~\ref{#1}}

\newcommand{\figref}[1]{Fig.~\ref{#1}}

\newcommand{\myparagraph}[1]{\vspace{0.05in}\noindent\textbf{#1}}
\newcommand*{\Cdot}{\raisebox{-0.25ex}{\scalebox{1.75}{$\cdot$}}}
\newcommand{\norm}[1]{\left\Vert#1\right\Vert}

\newboolean{draft-mode}
\setboolean{draft-mode}{true}
\newcommand{\sidenote}[1]{\ifthenelse{\boolean{draft-mode}}{\marginpar{\tiny\raggedright\textsf{\hspace{0pt}#1}}}{}}
\DeclareRobustCommand{\arnote}[1]{\ifthenelse{\boolean{draft-mode}}{\textcolor{blue}{\textbf{AR: #1}}}{}}
\DeclareRobustCommand{\ncdnote}[1]{\ifthenelse{\boolean{draft-mode}}{\textcolor{green}{\textbf{NCD: #1}}}{}}

\begin{document}

\title*{Sampling-based Planning of\\ In-Hand Manipulation with External Pushes}
\author{Nikhil Chavan-Dafle and Alberto Rodriguez}
\institute{Nikhil Chavan-Dafle and Alberto Rodriguez \at Department of Mechanical Engineering, MIT, USA\\
  \email{$<$nikhilcd,albertor$>$@mit.edu}}
%
%
\maketitle

\abstract{\hspace{1mm} This paper presents a sampling-based planning algorithm for in-hand manipulation of a grasped object using a series of external pushes. 
A high-level sampling-based planning framework, in tandem with a low-level inverse contact dynamics solver, effectively explores the space of continuous pushes with discrete pusher contact switch-overs.
We model the frictional interaction between gripper, grasped object, and pusher, by discretizing complex surface/line contacts into arrays of hard frictional point contacts.  The inverse dynamics problem of finding an instantaneous pusher motion that yields a desired instantaneous object motion takes the form of a mixed nonlinear complementarity problem.
Building upon this dynamics solver, our planner generates a sequence of pushes that steers the object to a goal grasp.
%
We evaluate the performance of the planner for the case of a parallel-jaw gripper manipulating different objects, both in simulation and with real experiments.
Through these examples, we highlight the important properties of the planner: respecting and exploiting the hybrid dynamics of contact sticking/sliding/rolling and a sense of efficiency with respect to discrete contact switch-overs.  
}


 \section{Introduction}
\label{sec:intro}

In-hand manipulation, understood as the capability to adapt a grasp on an object, facilitates the complex process involved in picking and using an object.
%
%
Robots, especially those with simple grippers, lack the necessary dexterity to do so, which strains their manipulation capabilities. 

\begin{figure}
    \centering
    \includegraphics[]{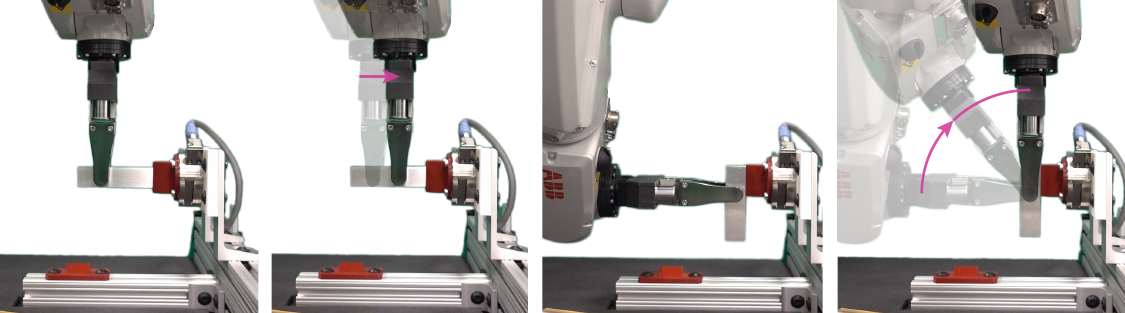}
    \caption{An example of prehensile pushing -- an aluminum object is reconfigured in a grasp by pushing it against the environment from different sides.}
    \label{fig:prepush_seq}
    \vspace{-0.2in}
\end{figure}

In this paper, we propose a planner to manipulate grasped objects through a sequence of external pushes, such as those in \figref{fig:prepush_seq}, a.k.a., prehensile pushing~\citep{ChavanDafle2015a}. 
Given a pair of start and goal object grasps, the planner outputs a sequence of pushes, possibly from different sides of the object, to reconfigure the object in the grasp.
Planning these push sequences presents two main challenges:
\begin{itemize}
    \item[$\Cdot$] \textbf{Continuous contact dynamics} of the frictional interaction between gripper, object and their environment.
    \item[$\Cdot$] \textbf{Discrete contact switch-overs} between continuous pushes.
\end{itemize}

To address them both, we combine a low-level optimization-based approach to solve the inverse dynamics of prehensile pushing with a high-level sampling-based planning approach to build long sequences of pushes. 

\myparagraph{Low-level optimization-based inverse dynamics}
For prehensile pushing, solving for a unit step control to propagate a planning-tree refers to solving the inverse dynamics problem, i.e., finding the external pusher motion that yields the object motion as close to the desired one as possible.
We develop an optimization-based dynamics formulation capturing the contact dynamics between gripper, object, and external pusher, which in practice takes the form of a mixed nonlinear complementarity problem (MNCP).
    
\myparagraph{High-level sampling-based planning}
The higher level planning architecture follows a transition-based RRT\mbox{*} (T-RRT\mbox{*}) formulation 
which takes advantage of the optimality convergence properties of typical RRT\mbox{*} technique and efficient exploration of configuration space using transition tests~\citep{trrt_star,trrt,rrt_star}. 
We use the optimal connections feature of RRT\mbox{*} to minimize the number of pusher contact switch-overs along a pushing strategy. 
The transition tests allow us to loosely confine the stochastic exploration towards the goal grasp, while allowing the flexibility to explore in other directions if it's necessary to get the object finally to the goal.
%

The planning architecture and the dynamics solver work together to build a tree of grasp poses. A path in this tree provides a pushing strategy to change a grasp pose to another.
%

We evaluate the performance of the planner for the case of a parallel jaw-gripper manipulating different objects. We validate the pushing sequences with real experiments in a robotic manipulation platform which is equipped to track the motion of the robot and pose of the object. 

To summarize, the main contributions of this paper are:
\begin{itemize}
    \item an optimization-based inverse dynamics formulation for full three dimensional in-hand manipulations using external pushes,
    
    \item a planning framework to combine low-level contact dynamics with high-level reasoning for long pushing strategies with discrete contact changes,
    
    \item application and experimental validation of the proposed planner to prehensile pushing
\end{itemize}

\section{Related Work}
\label{sec:related}

Early work on dexterous manipulation focused on providing a gripper with enough degrees of freedom to give full controllability over a grasped object and further allowing finger contacts to either roll or slide~\citep{salisbury1982ahf,Trinkle1990,Kao1992,Bicchi95a,Rus1999a,CherifGupta99}. It assumed the intrinsic capability of the gripper to control these interactions.

Diverging from this assumption, in a recent work, we demonstrated the use of gravity, dynamic motions, and contacts with the environment to regrasp objects using a library of hand-scripted motions~\citep{ChavanDafle2014}.
In~\citep{ChavanDafle2015a}, we studied in-hand manipulations with external contacts. We referred to it  as \textit{prehensile pushing} and presented a quasi-dynamic formulation to predict the instantaneous motion of a grasped object for a given pusher motion -- forward dynamics problem~\citep{ChavanDafle2015a}. The inverse dynamics solver we use in this paper shares a similar dynamics formulation underneath, but solves for the required pusher motion for a desired object motion.

Planning for prehensile pushing requires an understanding of how forces and motions evolve at contact interactions.
There is a large array of work on trajectory optimization techniques for planning and control through contact. In most cases, these make assumptions of point contact interactions modelled with polyhedral friction cones or patch contacts modelled as soft point contacts to alleviate the computational complexity of contact modeling \citep{Erez2012,todorov2012,Posa2014}.
\citet{lynch15}, \citet{kragic_pivoting2} and \citet{Hou16} demonstrate application of such approach to in-hand manipulation, particularly for in-hand sliding and pivoting. 
For computational efficiency, \citet{Erez2012,todorov2012} relax complementarity constraints required to impose non-penetration condition at contacts and to model sticking/sliding transitions.
This leads to fast algorithms, but with limited success in modeling situations of interest to this work, i.e., benefiting from hard line and patch contacts~\citep{kolbert16}.
The contact modelling approach in this paper resembles to that presented in~\citep{ChavanDafle2015a}, but with a quadratic Coulomb friction cone instead of a polyhedral approximation. The polyhedral approximation introduces artificial anisotropy in friction and ``preferred'' sliding directions.

Sampling-based techniques for planning are key to the presented approach.
Rapidly exploring random tree (RRT) derives it's strength from fast and random exploration of the configuration space~\cite{Lavalle98,Kuffner}. RRT\mbox{*} introduces the concept of optimality for connecting the nodes in a tree and provides conditions under which it can lead to asymptotically optimal solutions~\cite{rrt_star}.
One of the variants of RRT* that we particularly find useful in this work is T-RRT\mbox{*}, which is developed for path planning on configuration-space cost maps~\cite{trrt_star,trrt}. 
By employing a transition test to accept/reject nodes, it guides the exploration to follow low-cost valleys of the cost map with a provision to traverse across high-cost regions whenever required. This provides a more controlled and efficient exploration of the configuration space. 

While sampling based methods have not been thoroughly explored for contact-rich applications and may not seem an immediate choice for problems with complex and computationally expensive dynamics, in the coming sections we discuss in detail the fit of T-RRT\mbox{*} based approach for such problems and demonstrate its effectiveness at practical in-hand manipulations. 

\section{Problem Formulation}
\label{sec:formulation}

This paper focuses on planning in-hand manipulations using external pushes.
In our implementation, the external pushes are executed by a robot forcing a grasped object against a rigid environment.
More generally, such external pushes could also abstract the interactions with a second robot arm or extra fingers of a multi-finger gripper.

Equivalently, in this paper, we assume the gripper is fixed in the world and grasps an object, while a virtual pusher with full 6 DOF mobility executes the external pushes. In this case, planning for external pushes is equivalent to planning the motion of the virtual pusher. 
For the problem setup, we assume the following information about the manipulation system:
\begin{itemize}
    \item[$\Cdot$] Object geometry and mass.
    \item[$\Cdot$] Initial and goal grasp on the object, specified by the locations and geometries of each fingers contacts.
    \item[$\Cdot$] Gripping force.
    \item[$\Cdot$] Discrete set of pusher contacts, specified by initial locations and geometries.
    \item[$\Cdot$] Coefficient of friction at all contacts.
\end{itemize} 

As described in \secref{sec:intro}, the proposed planner works at two levels -- a high level planning architecture (\secref{sec:planning}) that explores the configuration space of reachable grasps and builds a tree of optimally connected configurations, and a low level inverse dynamics solver (\secref{sec:dynamics}) that controls the unit-step propagation in the tree.
In short, the decision flow of the planner is follows:
\begin{itemize}
    \item[i.] Sample a random object configuration in a grasp.
    \item[ii.] Check if moving toward the sampled configuration satisfies a ``benefit" criteria. If not, return to step i.
    \item[iii.] Solve inverse dynamics for a valid pusher location and pusher motion to move the object in the direction of the sampled pose. If not possible, return to step i.
    \item[iv.] Check for other ways to reach the newly added configuration with lower cost, from existing nodes in the tree.
    \item[v.] Iterate until reaching the goal grasp within a given resolution and cost threshold.
\end{itemize}

\section{Low-level: Inverse Dynamics Solver}
\label{sec:dynamics}

Sampling-based planners are built on top of a unit-step algorithm that, when possible, steers the system along a sampled direction.
In this paper, we refer to that unit step as the inverse dynamics problem:
given the pose of the object in a grasp, the position of a pusher on the object and a desired instantaneous object motion in the grasp, find an instantaneous motion of the pusher that forces the object in the direction as close to the desired direction as possible.

The following sections discuss our approach to model contact interactions and kinetic-kinematic constraints governing the object motion in the grasp.  

\subsection{Contact Modelling}
\label{sec:contact} 
\begin{wrapfigure}[5]{r}{2in}
\vspace{-15mm}
 \includegraphics[]{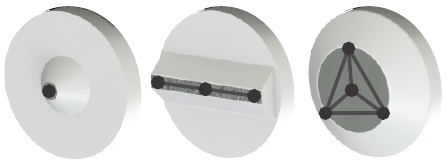}
\caption{Different contact geometries: point, line and circular patch, modeled as sets of rigidly connected point contacts}
    \label{fig:contact_shape}
\end{wrapfigure} 
Our contact modelling approach is similar to that proposed in~\citep{ChavanDafle2015a}. We model a patch contact as a rigid array of point contacts as shown in \figref{fig:contact_shape}.
Each of these constituent point contacts, is modeled as a hard point contact with quadratic Coulomb friction cone.

We represent a point contact between two bodies by a local coordinate frame with, $\hat{\boldsymbol{n}}$ normal to the contact plane and and $\hat{\boldsymbol{t}}$ and $\hat{\boldsymbol{o}}$ spanning the contact plane.
Let $\boldsymbol{f}=[f_{n}, f_{t},f_{o}]^\top$ and $\boldsymbol{v}=[v_n,v_t,v_o]^\top$ be a net force and a relative velocity at a contact in the local contact frame.
For a given coefficient of friction ($\mu$), Coulomb's friction cone at the contact is defined as the following set:
\begin{equation}
\label{eq:fricCone}
FC=\{f_{n}\hat{\boldsymbol{n}} +
f_{t}\hat{\boldsymbol{t}} + f_{o}\hat{\boldsymbol{o}}\
| \ f_{n} \geq 0,\ f_{t}^2+f_{o}^2 \leq \mu f_{n}^2\}
\end{equation}
By Coulomb's law, when a contact slides, the contact force is on the boundary of the friction cone and the direction of the friction force is opposite to that of the sliding velocity at the contact.
We can formalize this constraint using the standard complementarity and nonlinear equations:
\begin{equation}
\label{eq:compfricbound}
[(\mu f_n)^2 - f_t^2 -f_o^2] \norm{[v_t,v_o]} = 0, \ \ \ \  (\mu f_n)^2 - f_t^2 -f_o^2\geq 0
\end{equation}
\vspace{-3mm}
\begin{equation}
\label{eq:quad_dissipation}
  \mu f_{n}v_{i}+f_{i}\norm{[v_t,v_o]}=0 \hspace{1cm} i=t, o
\end{equation}
%

%
For a contact with finite area, modelled as an array of points, we impose \eref{eq:compfricbound} and \eref{eq:quad_dissipation} at each constituent point, along with constraints on the relative velocities at them to make sure that the array moves as a rigid body. See~\cite{ChavanDafle2015a} for more details.

\subsection{Dynamics of Prehensile Pushing}
\label{sec:constraints}
%
The fictional forces involved in prehensile pushing are much more dominant than the object inertia, so we will limit ourselves to a quasi-dynamic model of pushing. We define this model in the space of local contact impulses, relative velocities at the contacts, velocity of the object, and velocity of the pusher. The solution space is constrained by the following kinematic and kinetic constraints.

\myparagraph{Newton Euler Equation}:
\label{sec:newton}
Let $\mathbf{G_i}$ maps local contact forces at contact $i$ to the corresponding wrench in the object frame. $\mathbf{G}$ is defined as diagonal concatenation of $\mathbf{G_i}$'s for all the contacts on the object. 
As we are interested in a quasi-dynamic formulation, for a single time step with zero initial velocity of the object, we can write the time-integrated Newton's law for an object with mass $m$ and generalized inertia matrix $\mathbf{M}$ as:
\begin{equation}
\label{eq:newton_vel}
\mathbf{G}\cdot \boldsymbol{P} + \vec{P}_{mg} = \mathbf{M}\cdot {\vec{v}}_{\textnormal{obj}}
\end{equation}

\noindent where $\boldsymbol{P}$ is an array collecting impulses equivalent to all the contact forces ($\boldsymbol{f_1}, . ., \boldsymbol{f_n}$), $\vec{P}_{mg}$ is the gravitational impulse and ${\vec{v}}_{\textnormal{obj}}$ is the resultant object velocity in the object frame.

\myparagraph{Rigid Body Motion Constraints}:
\label{sec:rigid_body}
Let $\mathbf{J}$ be the jacobian matrix that maps the velocities of the pusher and gripper actuators ($\dot{\theta}$) to the input velocities at all the contacts in the local contact frames. We can write $\boldsymbol{V} = [\boldsymbol{v_1}, \boldsymbol{v_2}, . . ., \boldsymbol{v_n}]^\top$, the array collecting the relative velocities at all contacts, as difference between the input velocities and the reflection of the object velocity at those contacts points:
\begin{equation}
\label{eq:contact_vel}
\boldsymbol{V} = \mathbf{G}^\top\cdot\vec{v}_{\textnormal{obj}} - \mathbf{J}\cdot \dot{\theta}
\end{equation}

\myparagraph{Unilateral Contact Constraints}:
\label{sec:unilateral}
There can not be interpenetration at contacts between two rigid bodies. Contacts can only push and not pull, and only when there in no separation at them. We write it as a complementarity constraint at each point contact.
\begin{equation}
\label{eq:unilateral2}
  v_{n}\cdot p_{n}=0,\  v_{n}\geq 0,\
  p_{n}\geq 0
\end{equation}

\myparagraph{Contact Modelling Constraints}:
We model the force-motion interactions at every contact as explained in \secref{sec:contact}. Let $\boldsymbol{p}=[p_{n},p_{t},p_{o}]^\top$ be an impulse at a contact, then rewriting equations \ref{eq:compfricbound} and \ref{eq:quad_dissipation} in the space of impulse-velocity:
\begin{equation}
\label{eq:compfricbound2}
[(\mu p_n)^2 - p_t^2 -p_o^2] \norm{[v_t,v_o]} = 0, \ \ \ \ (\mu p_n)^2 - p_t^2 -p_o^2\geq 0
\end{equation}
\begin{equation}
\label{eq:quad_dissipation2}
  \mu p_{n}v_{i}+p_{i}\norm{[v_t,v_o]}=0 \hspace{1cm} i=t, o
\end{equation}
Further, for contacts with finite area modelled with arrays of point contacts, we impose constraints on the relative velocities at them to make sure that each array moves as a rigid body.

\subsection{Numerical Solver for the Dynamics Problem}
\label{sec:dyn_solving}

In our problem, solving inverse dynamics means finding a pusher velocity that produces a desired object velocity while satisfying all the constraints listed above. It has the form of a mixed nonlinear complementarity problem (MNCP), which we solve as a nonlinear constrained optimization problem using interior point method in MATLAB. We define the objective function as a weighted sum of the complementarity constraints and the difference between the desired object velocity and that actually achieved. We try to minimize the objective function subject to the constraints detailed in Section \ref{sec:constraints}. A feasible solution exists when the objective goes close to zero while meeting the constraints. 
In practice, it helps to give a relatively larger weight on complimentarity constraints, yielding more accurate satisfaction of contact dynamics and compromising on the desired object velocity if necessary. The ratio of weights we used is $10^4$.



\section{High-Level: Long Horizon Planning with Contact Switch-overs}
\label{sec:planning}
An effective regrasp skill requires exploiting  contact switch-overs. A continuous and greedy approach based on pushing iteratively towards the goal grasp has limited success in a problem as constrained and underactuated as in-hand manipulation.
The problem benefits from a long-horizon planning technique that allows the regrasp strategy to deviate from goal momentarily if necessary and sequence different discrete pushes.
%
%

Trajectory optimization has been studied to capture the effects of a long-horizon cost, but has difficulty with the hybridness of discrete contact switch-overs.
On the other hand, sampling based methods are naturally suited to search over continuous plans intertwined with discrete changes along the plan.
Being able to change the pusher contact from one side of the object to another can be pivotal. In practice, minimizing the number of contact switch-overs yields benefits in the form of: time savings, and avoiding uncertainty introduced by engaging and disengaging contacts.

%
The higher level architecture of our planner is based on a T-RRT\mbox{*} formulation. 
We exploit the optimality convergence properties of the underlying RRT\mbox{*} method to reduce the number of pusher contact switch-overs and the efficiency of transition tests to direct the exploration of configuration space towards the goal. 

\begin{algorithm}
  \caption{: In-Hand Manipulation Planner}\label{alg:full_planner}
  $  \textbf{input}: q_{init}, q_{goal}$ \par
  $  \textbf{output}:$ {tree} $\ \mathcal{T}$
  \begin{algorithmic}[1]
  \State $\mathcal{T}\gets \textrm{initialize tree}(q_{init})$
      \While{$q_{goal} \notin \mathcal{T}$}
        \State $q_{rand}\gets \textrm{sample random configuration}(\mathcal{C})$
        
        \State $q_{parent}\gets \textrm{find nearest neighbor}(\mathcal{T},q_{rand})$
        
        \State $q_{ideal}\gets \textrm{take unit step}(q_{parent},q_{rand})$
        
        \If{\textrm{transition test}$(q_{parent},q_{ideal},\mathcal{T})$ \textbf{and} \textrm{grasp maintained}$(q_{ideal})$} 
            
            \State $q_{new}, \dot{\theta}_{pusher} \gets \textrm{InvDynamics}(q_{parent},q_{rand})$
            
            \If{$q_{new}\neq \textrm{null}$ \textbf{and} \textrm{transition test}$(q_{parent},q_{new},\mathcal{T})$ \textbf{and} \textrm{grasp maintained}$(q_{new})$}
            \State $(q\mbox{*}_{parent})\gets \textrm{optimal connection}(\mathcal{T},q_{new},q_{parent})$
            
            \State $\textrm{add new node}(\mathcal{T},q_{new})$
            
            \State $\textrm{add new edge}(q\mbox{*}_{parent},q_{new})$
            
            \State $\textrm{rewire tree}(\mathcal{T},q_{new},q\mbox{*}_{parent})$
            
            \EndIf
        \EndIf
    \EndWhile
  \end{algorithmic}
\end{algorithm}

Algorithm~\ref{alg:full_planner} presents our in-hand manipulation planner, starting from the assumptions listed in \secref{sec:formulation}. 
Let $q$ denote a configuration of an object, i.e., the pose of the object with respect to a gripper frame which is assumed to be fixed in the world. Though the configuration space $\mathcal{C}$ is six dimensional, different types of grasps confine it to lower dimensions.
We use scaled euclidean distance, where $1$~mm is treated equivalent to $3$~degree, as a metric between two object configurations.

Let $q_{init}$ and $q_{goal}$ be an initial and desired configuration of the object respectively. 
The planner initiates a tree $\mathcal{T}$ with $q_{init}$. 
While the desired object pose is not reached, it samples a random configuration ($q_{rand}$) and finds the nearest configuration to $q_{rand}$ in the tree $\mathcal{T}$.

\myparagraph{Controlled exploration}: A transition test decides if the propagation of the tree towards the newly sampled configuration is acceptable or not.
Let $C_{q}$ be a cost defined on the object configuration $q$, as the distance between $q$ and  $q_{goal}$.
If moving the object from the nearest neighbor towards the newly sampled pose can reduce the configuration cost, the sample is accepted.
If such an object motion will increase the configuration cost, but still keep it lower than some maximum bound set, the sample is accepted with a certain probability. Following \citet{trrt}, we define the transition probability for a transition from $q_a$ to $q_b$ as:

\centerline{$p_{(q_a,q_b)} =\exp(\frac{-\Delta C_{(q_a,q_b)}}{KT})$}

\noindent where,
\begin{itemize}
    \item[$\Cdot$] $\Delta C_{(q_a,q_b)} = \frac{C_{q_b}-C_{q_a}}{dist(q_a,q_b)}$ is the rate of cost variation per unit distance.
    \item[$\Cdot$] $K$ is a normalization factor defined as average of costs $C_{q_b}$ and $C_{q_a}$.
    \item[$\Cdot$] $T$, is temperature parameter which controls the difficulty of a transition. We adjust it as the planner progresses. It is increased if the tree is getting stuck locally to allow transitions of high cost, and decreased otherwise to allow transitions of only low cost. 
\end{itemize}

In practice, the transition test with our configuration cost definition loosely confines the propagation of the tree towards the goal pose, while allowing the flexibility to steer away from it momentarily if necessary.

If the transition test succeeds, we query the inverse dynamics solver to predict the motion of a pusher required to move the object from ($q_{parent}$) by a unit step as much as possible towards ($q_{rand}$). 
The dynamics solver limits its choice for the pushers to a fixed set of pusher locations on the object and the evolved pusher location corresponding to $q_{parent}$. Here, by evolution we mean the new location of the pusher contact if it slides on the object.
This makes sure that we account for the pusher slip when sequencing multiple instantaneous pushes to generate a smooth continuous push using the same pusher.

\vspace{-5mm}
\noindent
\begin{minipage}[t]{5.85cm}
\null 
 \begin{algorithm}[H]
    \caption{: optimal connection}\label{alg:optimConnect}
   $  \textbf{input}: \mathcal{T}, \ q_{new}, \ q_{parent} $ \par
  $  \textbf{output}: q\mbox{*}_{parent}$
    \begin{algorithmic}[1]
  \State $J_{q_{new}}\gets \textrm{findNodeCost}(q_{new},q_{parent})$
  \State $J\mbox{*}_{q_{new}}\gets J_{q_{new}}$; \ \  $q\mbox{*}_{parent}\gets q_{parent}$
  \State $Q_{near}\gets \textrm{nodesInBall}(\mathcal{T},q_{new},R_{ball})$
  
  \While {$Q_{near} \neq \emptyset$} \par
  \State $q_{parent}\gets q \in Q_{near}$
  \If{$\textrm{InvDynamics}(q_{parent},q_{new})\neq\textrm{null}$}
  \State $J_{q_{new}}\gets \textrm{findNodeCost}(q_{new},q_{parent})$
  
  \If{$J_{q_{new}}<J\mbox{*}_{q_{new}}$}  \par
    \State $J\mbox{*}_{q_{new}}\gets J_{q_{new}}$
    \State $q\mbox{*}_{parent}\gets q_{parent}$  
  \EndIf
  \EndIf
  \State $Q_{near}\gets Q_{near} \setminus q_{parent}$
  \EndWhile  
  \end{algorithmic}
  \end{algorithm}
\end{minipage}%
\begin{minipage}[t]{5.75cm}
\null
 \begin{algorithm}[H]
    \caption{: rewire tree}\label{alg:rewireTree}
 $  \textbf{input}: \mathcal{T}, \ q\mbox{*}_{parent}, \ q_{new}$ \par
  $  \textbf{output}: \textrm{tree} \ \mathcal{T}$
    \begin{algorithmic}[1]
  
  \State $q_{parent}\gets q_{new}$
  \State $Q_{near}\gets \textrm{nodesInBall}(\mathcal{T},q_{new},R_{ball})$
  
  \While {$Q_{near} \neq \emptyset$} \par
  \State $q_{r}\gets q \in Q_{near}$
  \State $q_{parent}\gets q_r.{parent}$
  \State $J_{qr} \gets \textrm{findNodeCost}(q_{r},q_{parent})$
  \If{$\textrm{InvDynamics}(q_{new},q_{r})\neq\textrm{null}$}
  \State $J_{qr_{new}}\gets \textrm{findNodeCost}(q_{r},q_{new})$
  
  \If{$J_{qr_{new}}<J_{qr}$}  \par
    \State $q_r.{parent}\gets q_{new}$  
  \EndIf
  \EndIf
  \State $Q_{near}\gets Q_{near} \setminus q_{r}$
  \EndWhile
  \end{algorithmic}
  \end{algorithm}
\end{minipage}
\vspace{4mm}

\myparagraph{Optimal connections}: As we wish to minimize the number of pusher contact switch-overs, we define the cost of a node in a tree ($J_q$) to reflect the contact switch-overs performed to get to that node from the start node. Formally,

\centerline{$J_q = J_{q_{parent}} + \ dist(q,q_{goal}) \ + \ $cost of the instantaneous push.}

\noindent where, the cost of the instantaneous push that would move the object from $q_{parent}$ to $q$, is set to 0.1 (low) if the pusher used to get to $q_{parent}$ is used in continuation for this instantaneous push, or 1 (high) if the pusher location is changed. For reference, distance from the goal is generally in the order of $10^{-3}$ to $10^{-1}$.

Using this node cost definition, \textit{optimal connection} routine explores the space around $q_{new}$ to find transitions that lead to a lower cost for $q_{new}$, and iteratively updates the parent node of $q_{new}$ and the cost of $q_{new}$ accordingly.  
Similarly, \textit{rewire tree} routine checks if any of the nodes around $q_{new}$ can be connected through $q_{new}$ with the purpose of reducing its cost. Both these routines are characteristics of the RRT* architecture originally proposed in~\cite{rrt_star}. 

To summarize, the high level planner generates a tree of grasp poses connected with continuous pushes or with discrete pusher switch-overs. A path in this tree is a long pushing sequence that changes the grasp on the object from one pose to another with a small number of pusher contact switch-overs if necessary.

\section{Example Cases}
\label{sec:examples}
In this section, we consider different examples of in-hand manipulation while highlighting notable features of the planner. For all the experiments, we used a computer with Intel Core i7 2.8 GHz processor and MATLAB R2016a. We evaluate the validity of the solutions with a manipulation platform instrumented with an industrial robot arm, a parallel-jaw gripper with force control at the fingers, features in the environment that will act as pushers, and a Vicon system for object tracking.

\vspace{-1mm}
\subsection{Respect and Exploit Dynamics of Frictional Contact}
\label{sec:linpush}
\vspace{-7mm}
\begin{figure}
    \centering
    \includegraphics[width=4.5in]{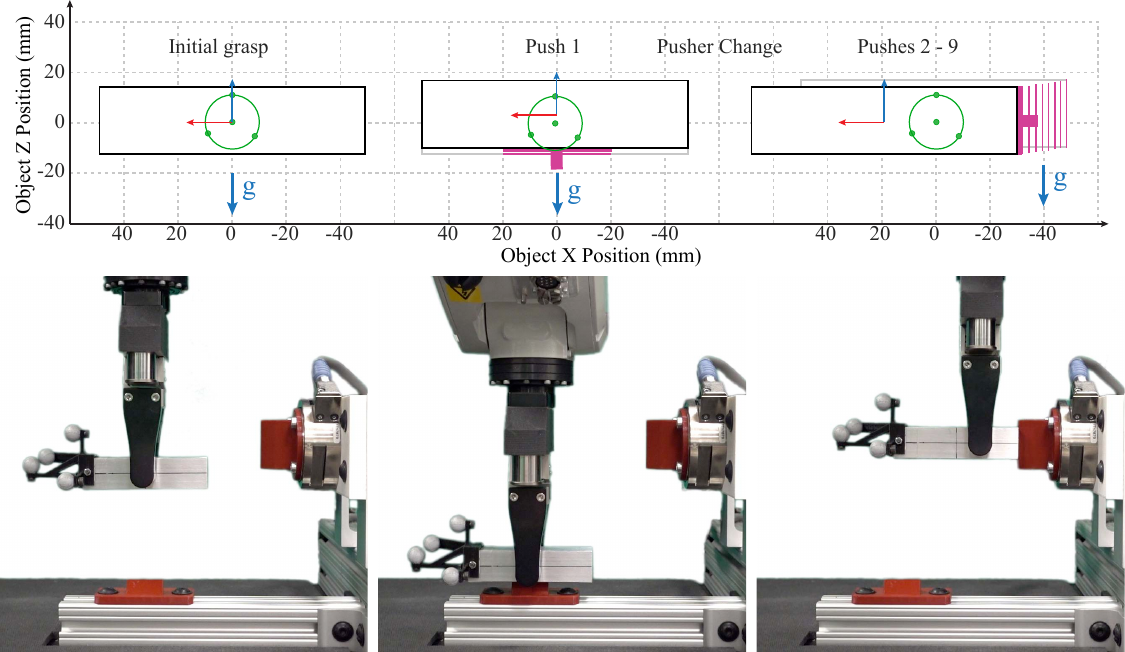}
    \caption{Simulated motion of the object and snapshots from the experiment for a pushing sequence generated by the planner. Object motion is shown from a side view; finger contact is a circular patch (shown in green) and pusher contact is a line/edge contact (shown in magenta).}
    \label{fig:linpush_simexp}
\end{figure}
\vspace{-10mm}
\begin{SCfigure}
{\includegraphics[]{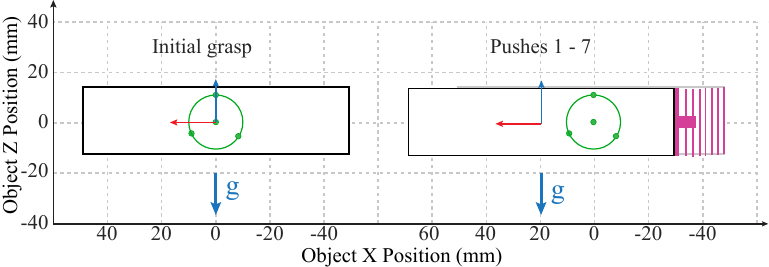}}
{\caption{Simulated motion of the object for a pushing sequence for a light-weight plastic object. Note that only side pusher is used throughout and downward sliding of the object is minimal.}
\label{fig:plastic_push}}
\end{SCfigure}

Having a detailed underlying dynamics solver is one of the key strengths of our planner. This example shows different strategies generated by the planner to execute the same manipulation for two similar objects, but of different weights.

\begin{figure}
    \centering
    \includegraphics[width=4.5in]{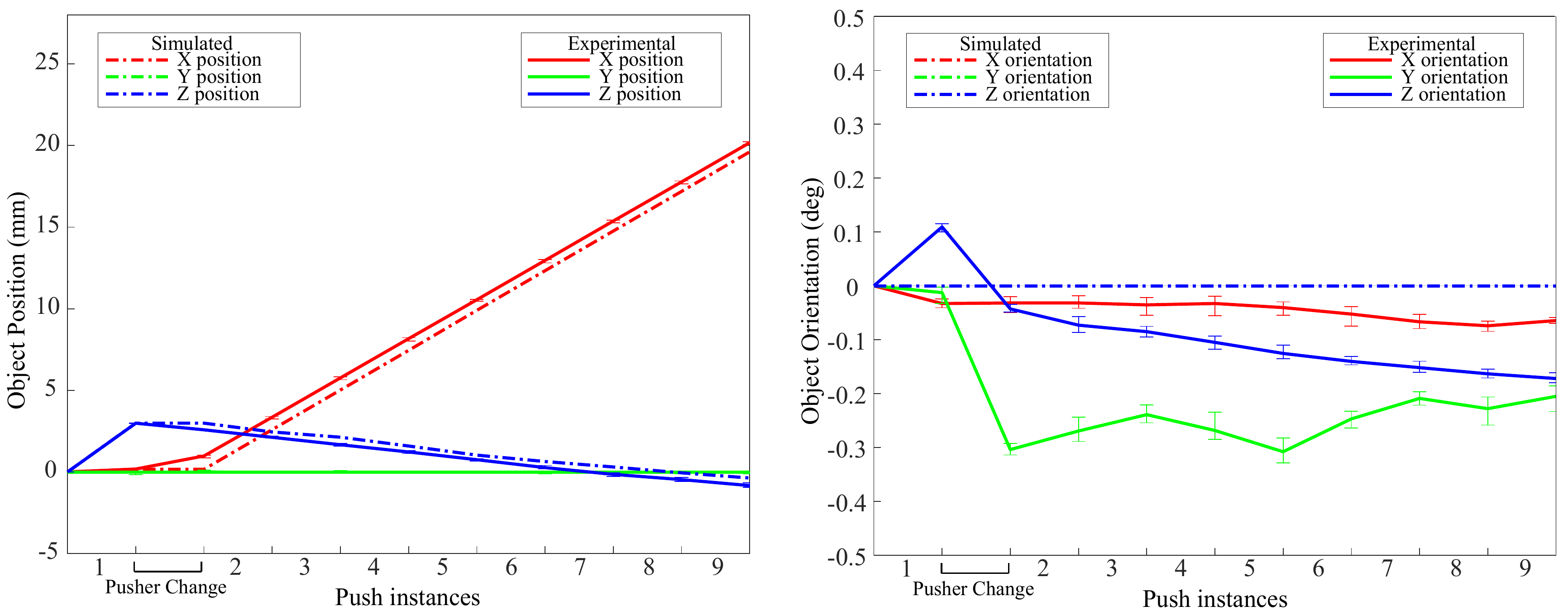}
    \caption{Object motion in the grasp as predicated by the planner and as observed in the experiment for the example shown in \figref{fig:linpush_simexp}. Mean values for 10 experimental runs are shown, with error-bars indicating the variation observed during these runs.}
    \label{fig:linpush_plot}
    \vspace{-4mm}
\end{figure}

First consider a $100$~mm long aluminum bar of $1$~inch square cross-section grasped at its center with a parallel-jaw gripper. The goal in this seemingly simple manipulation is to move the object to a pose $20$~mm offset in the horizontal direction from the center.
The combination of the coefficient of friction at the fingers and the pusher and the gripping force make it so that the downward sliding of the object under gravity is not negligible. Pushing the object horizontally from side is not a valid solution.

We initiate the planner with pusher contacts on left, right, and bottom face of the object. Note that in all the examples we consider in this paper, the robot is constrained to use features in the environment as virtual pushers, so the gravity direction remains constant in the pusher frame and is different in different contact frames based on their orientation in the environment. 
\figref{fig:linpush_simexp} shows a pushing sequence generated by the planner and consequent motion of the object in the grasp. The object is pushed up first using the bottom pusher. This helps to account for the downward sliding of the object due to gravity in the later pushes from side. Note that the planner decided to do this upward push first, even though it means going away from the goal pose; this strategy leads to only one pusher switch-over in the process of getting object to the goal pose. The median time taken to converge to a plan with only one pusher switch-over for 10 trials was 9.88 minutes.

Now, consider the same problem but for a plastic object which weighs half of the aluminum object and has similar frictional properties. For this case, the planner decides to push only from side, as shown in \figref{fig:plastic_push}. The downward sliding of the object during these pushes is minimal and the final object pose in the grasp is within the desired resolution from the goal pose.
Experimentally, the plastic object indeed slides down by a negligible amount and we get the horizontal displacement of the object in the grasp as desired.

%
\figref{fig:linpush_plot} shows the comparison between the object motion in the grasp simulated by the planner and that observed during experimental trials for the aluminum object.
We get about $0.56$mm error in X and $0.45$mm error in Z in the final position of the object in the grasp from what is expected by the planner. The errors in the orientation are less than $0.25$ degree.
Due to high precision of the robot, the experiments are very repeatable and the error-bars in \figref{fig:linpush_plot} showing the variation in 10 experiments are almost non-visible in the position plot.

\subsection{Minimize the Number of Contact Switch-overs}
\label{sec:pivoting}
\vspace{-6mm}
\begin{figure}
    \centering
    \includegraphics[width=4.5in]{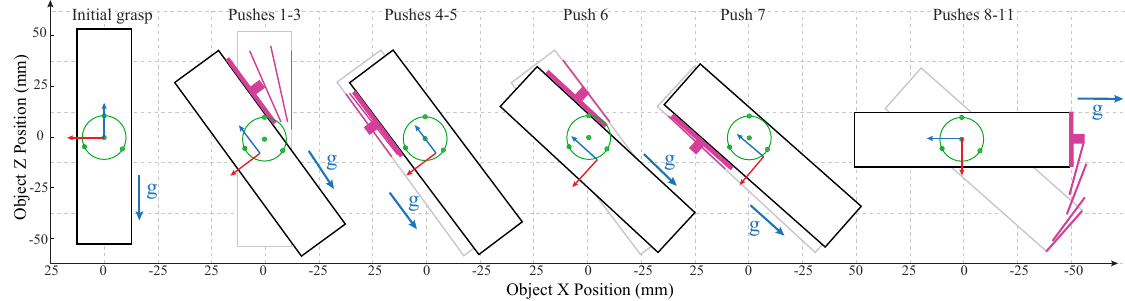}
    \caption{A pushing sequence for pivoting the aluminum object in a parallel-jaw grasp. The pushing sequence involves discrete pusher switch-overs to push the object from different facets to eventually get to the desired pose. 
    }
    \label{fig:pivoting_okseq}
\end{figure}
\vspace{-10mm}
\begin{figure}
    \centering
    \includegraphics[width=4.5in]{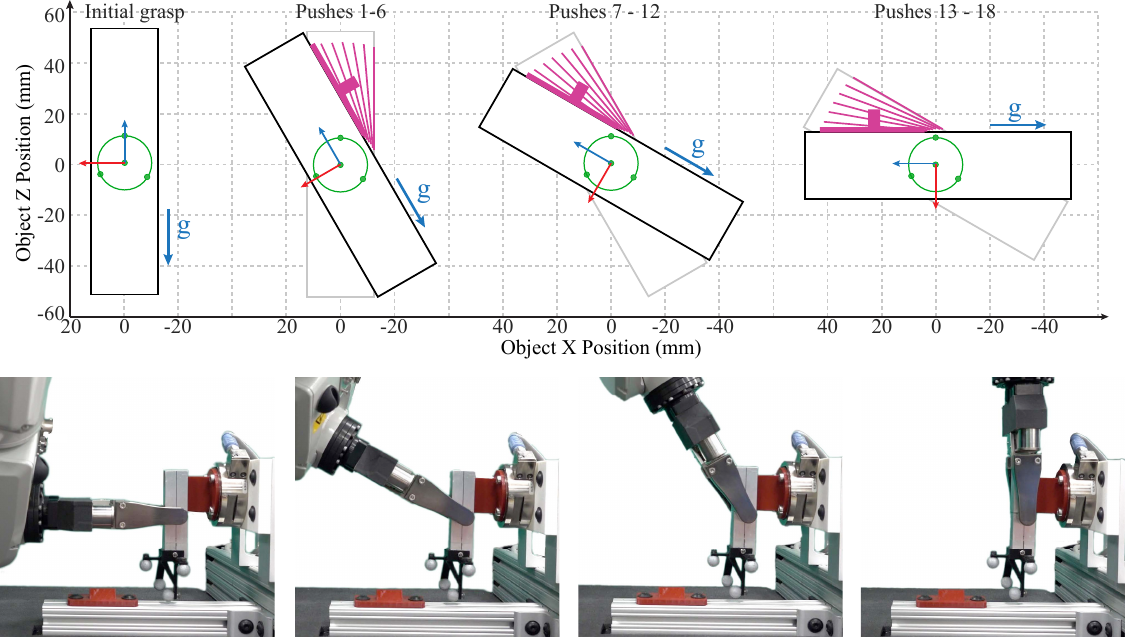}
    \caption{Pivoting strategy generated using a single pusher contact on right face of the object. 
    }
    \label{fig:pivoting}
    \vspace{-2mm}
\end{figure}

In this example, the goal is to pivot the same aluminum object about the fingertips by 90 degrees. We initiate the planner with pushers on left, right and bottom face of the object. \figref{fig:pivoting_okseq} shows a series of pushes and consequent object motion generated by the planner. Note that the planner uses all three contacts to eventually pivot the object by 90 degree and to correct the unwanted object displacements happened during those pushes. 

\begin{figure}
    \centering
    \includegraphics[width=4.5in]{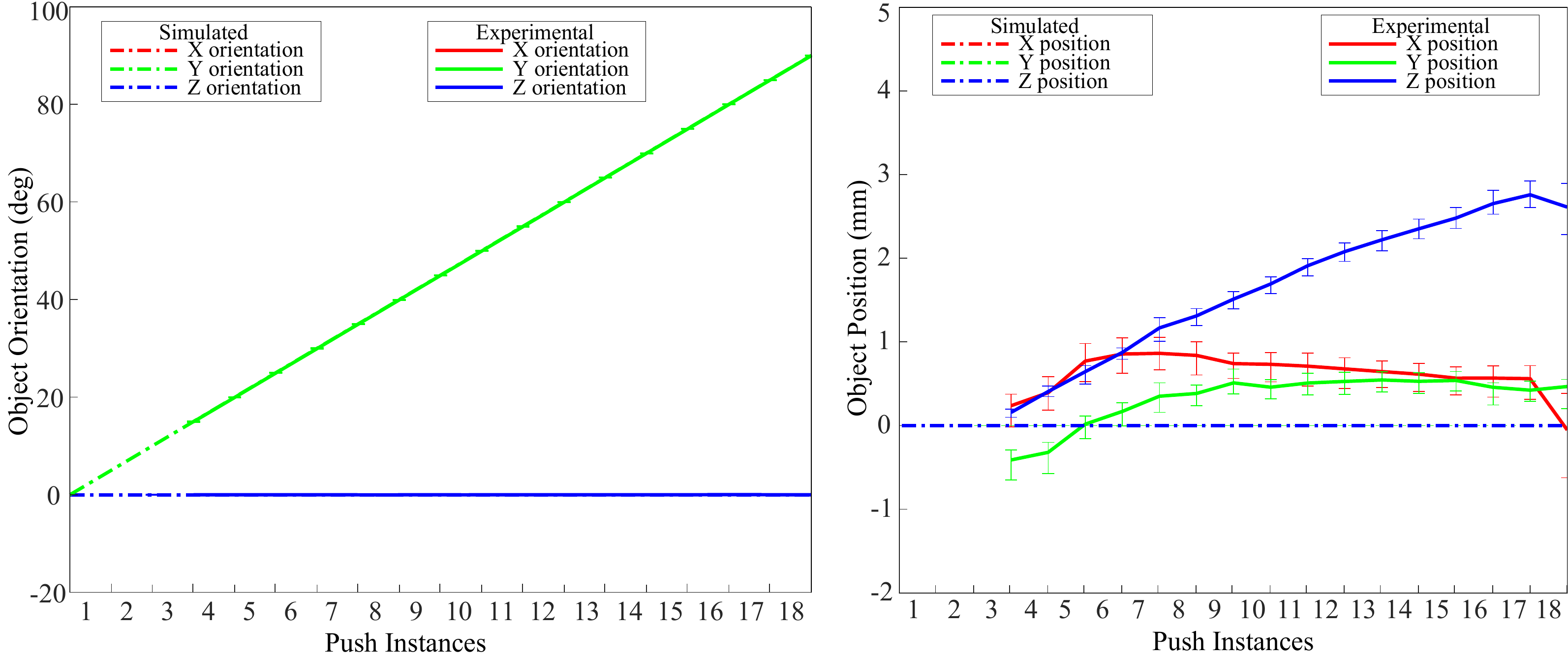}
    \caption{Object motion in the grasp as predicated by the planner and as observed in the experiment for the example shown in \figref{fig:pivoting}. Mean values for 10 experimental runs are shown, with error-bars indicating the variation observed during these runs.}
    \label{fig:pivot_plot}
    \vspace{-4mm}
\end{figure}

In another attempt, we introduce a bias in the definition of the distance metric used to find the nearest node for connection. We influence the distance metric more by the difference in the position than that in the orientation. This promotes the connections between the object poses that are close in terms of positions but may have different orientations.
The planner converges to a pivoting strategy in which a single pusher rotates about the fingertips to pivot the object with almost no object displacement in the grasp. \figref{fig:pivoting} shows instances of pushing strategy generated by the planner and corresponding snapshots of the experimental run.
Note that the gravity is constant in the pusher frame as shown in \figref{fig:pivoting}.
The median time taken to generate this plan for 10 different attempts was $2.14$ minutes.

This example shows that with our TRRT*-based formulation and node cost definition a pushing strategy converges to the one with fewer number of pusher changes, and providing some heuristic can further speed up that process.

\figref{fig:pivot_plot} shows comparison between the simulated object motion and that observed experimentally for the plan shown in \figref{fig:pivoting}. Error-bars show the variations during 10 experimental runs. For first two pushes, the Vicon markers on the object get occluded by the robot. So, the experimental values are shown in the plots only after the third push.
We find a close match between the orientation of the object as predicted by the planner and that seen during the experiments; the object position however shows some deviations.
Final position of the object in the grasp is moved along Z by $2.5$mm and in Y by $0.5$mm which the planner does not expect.

The errors observed in this as well as the previous example can be attributed to a few possible sources such as the errors in locating the pusher contacts in the environment, unmodeled compliance of the fingers and gripper mechanism, and possible manufacturing defects in the finger and pusher contacts.

\subsection{Exploit Complex Contact Interactions}
\label{sec:ballrolling}
\begin{figure}
    \centering
    \includegraphics[width=4.5in]{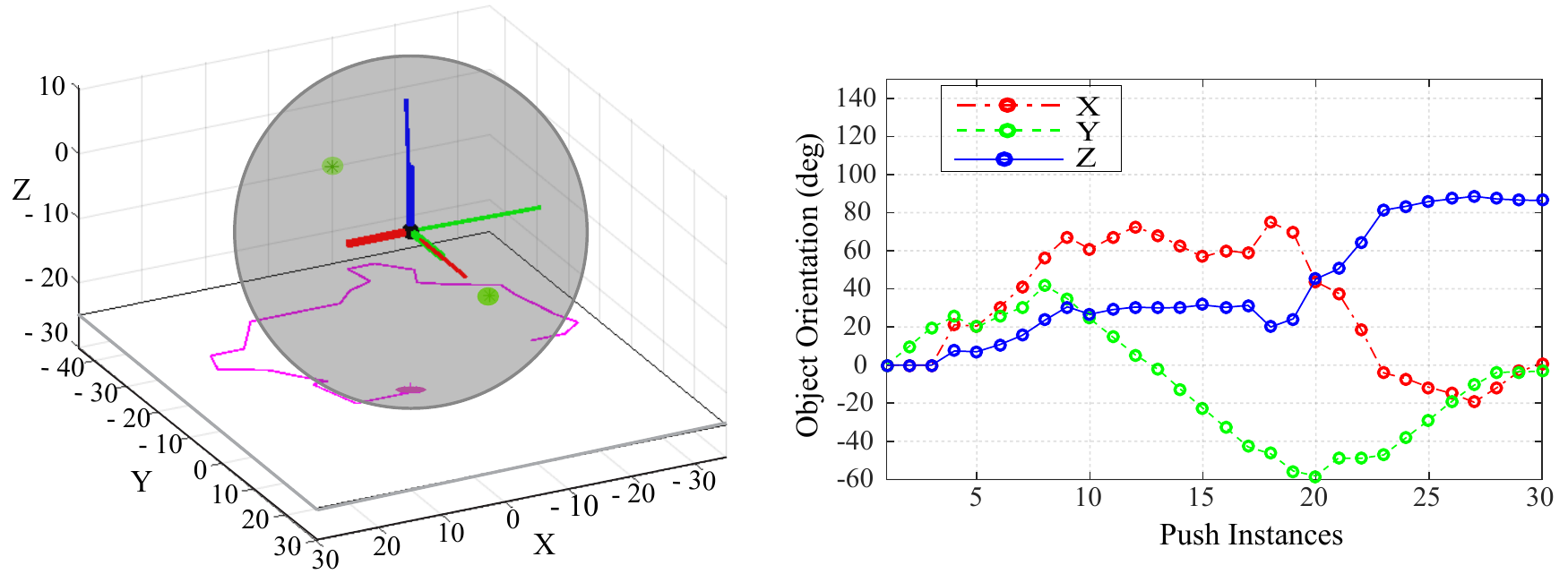}
    \caption{Evolution of the rolling contact and orientation of the ball in the grasp for the trajectory planned to rotate the ball about Z axis. Finger contacts are shown in green, while the contact between the ball and the ground is shown in magenta color.}
    \label{fig:rolling_sim}
\end{figure}
\begin{figure}
    \centering
    \vspace{-1mm}
    \includegraphics[width=4.5in]{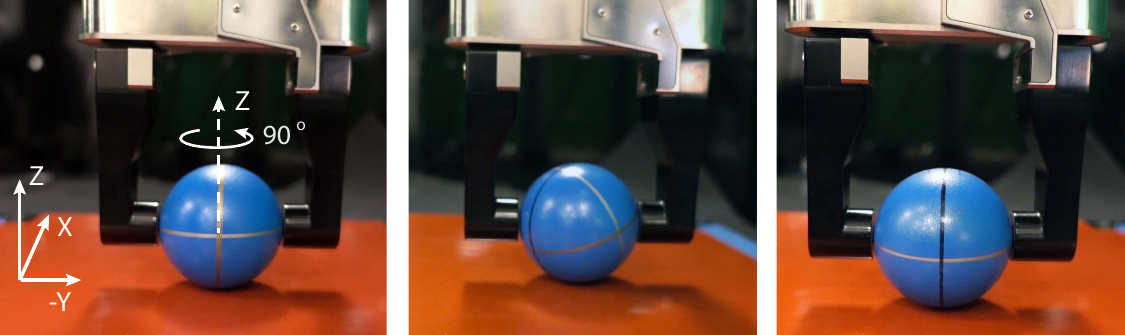}
    \caption{Object pose in the grasp at the beginning, middle and end of the rolling trajectory. Black, silver and golden paint marks on the ball, show that the object effectively rotates by 90 degrees about vertical (Z) while net orientation about other two axes (X and Y) go close to zero as before. The supplemental video shows the actions involved better.}
    \label{fig:rolling_exp}
    \vspace{-3mm}
\end{figure}
This example is similar to the classical ball-plate problem \citep{ball_plate}. Imagine a steel ball in a parallel-jaw grasp and resting on a ground as shown in \figref{fig:rolling_exp}. We wish to rotate the ball in the grasp about vertical (Z axis) by 90 degree using the ground as a virtual pusher. As the contact between the ball and the flat ground is of very small area, theoretically a point contact, it can not rotate the ball about Z using friction.
When provided with this challenge, the planner generates a series of in-plane pushes that causes the ball to purely rotate about X and Y axes in the grasp and eventually go to the orientation with close to 90 degree rotation about Z and almost zero rotations about X and Y.
The time taken to generate this plan was 318.17 minutes.

\figref{fig:rolling_sim} shows the rolling contact trajectory of the ball and the orientation the ball along it.
Note that the ball is free to rotate about the axis connecting fingers (Y axis) as the finger contacts are point contacts; however, rotation about X needs to overcome friction and locally slide at fingers along the vertical direction (Z).
All the contacts are free to stick or slip. For the planned trajectory, the contact between the ground and the object is instantaneously sticking, i.e. rolling contact, while there is sliding at the fingers contacts only in the vertical direction (Z) to allow the ball to rotate about X axis with no change in the position of the ball in the grasp.

Realizing such rolling in the grasp is easier when either the gripping force is very low or the coefficient of friction at the pusher contact is much higher than that at the fingers. We use a high friction silicone platform as a ground pusher.
Since we did not have a way to track the pose of the ball accurately, we provide only qualitative results for this example.
\figref{fig:rolling_exp} shows the snapshots of the actual implementation of the ball rolling example on our system. It shows rotation of the ball by 90 degrees about the vertical axis. The rotation about the other axes is close to zero and the object position in the grasp remains intact.

\section{Discussion}
\label{sec:discussion}

This paper presents a sampling-based planning framework for in-hand manipulations using external pushes. 


We model the frictional interactions between the grasped object, fingers, and the environment with a quadratic Coulomb friction cone and complementarity constraints capturing the hybrid nature of sticking/sliding. The resulting inverse dynamics problem for estimating the pusher velocity to produce a desired instantaneous object velocity in the grasp naturally takes the form of MNCP and is solved as a nonlinear constrained optimization problem.

The high-level planning architecture is based on T-RRT\mbox{*} and relies on the inverse dynamic model of prehensile pushing as the underlying unit-step controller to propagate states. We exploit the strengths of T-RRT\mbox{*} for two specific purposes:
1) to bias the exploration towards the goal pose with a provision to deviate from the goal whenever necessary, and 2) to build low-cost connections in the tree that yield effective pushing strategies for regrasps while avoiding unnecessary pusher contact switch-overs.

We evaluate the planner with a parallel-jaw gripper manipulating different objects.
Simulation results show that our planning framework is able to exploit the dynamics of pushing and reason about strategies with continuous pushes linked with discrete pusher contact switch-overs. The experimental observations validate the accuracy of the generated plans; the planned strategies move the object very close to the desired pose in the grasp.

The main limitations of the current approach are:
\begin{itemize}
\item \textbf{Speed} The inverse dynamics formulation we developed is computationally expensive which consequently affects the planning time. This is inline with existing algorithms that use complimentarity formulations to explicitly model the hybrid dynamics of rigid contact~\cite{Posa2014,tassa2010stochastic}. This work focuses on demonstrating the effective blend of a detailed dynamics modelling and a sampling-based method for planning in-hand manipulations. It is entirely developed in MATLAB for flexibility and currently not optimized for time. 

One promising direction for faster planning is to limit the planner to a subset of pushing motions whose dynamics are less expensive to compute~\cite{ChavanDafle2018a}. Another practical way is to extend this work to a multi-query framework to exploit the already built tree/graph. This can work better for applications such as assembly automation where robots often deal with a small set of known objects, initial grasps and goal grasps.

\item \textbf{Smoothness} The solutions tend to be jerky, as it is typical from randomized sampling-based planners. It would be interesting to explore the role that trajectory optimization approaches can play in bolstering sampling-based methods.
\end{itemize}

An approach to in-hand manipulation that is not limited to intrinsic dexterity, but relies on external contacts to produce the desired reconfigurations can make robots more flexible and reliable at autonomous manipulation, even those robots with simple grippers currently involved in today's factories and field applications.

\medskip

\small{\bibliography{ncd-isrr17}}

\end{document}